\documentclass[10pt,twocolumn,letterpaper]{article}

\usepackage{wacv}
\usepackage{times}
\usepackage{epsfig}
\usepackage{graphicx}
\usepackage{amsmath}
\usepackage{amssymb}
\usepackage[includeheadfoot,margin=2cm]{geometry}
\usepackage[font=small,labelfont=bf,tableposition=top]{caption}
\usepackage{footnote}
\usepackage{subfigure} 
\usepackage{amsthm} 
\usepackage{listings}
\usepackage[numbers,sort&compress]{natbib}

\newcommand{\cutabstractup}{\vspace*{-0.15in}}

\newcommand{\cutsectionup}{\vspace*{-0.1in}}
\newcommand{\cutsectiondown}{\vspace*{-0.01in}}

\newcommand{\cutsubsectionup}{\vspace*{-0.07in}} 
\newcommand{\cutsubsectiondown}{\vspace*{-0.05in}} 

\newcommand{\cutsubsubsectionup}{\vspace*{-0.1in}} 
\newcommand{\cutsubsubsectiondown}{\vspace*{-0.05in}} 

\makeatletter
\newcommand{\thickhline}{%
    \noalign {\ifnum 0=`}\fi \hrule height 1.0pt
    \futurelet \reserved@a \@xhline
}

\newcommand{\mvg}{\text{multivariate Gaussian}}

\newcommand{\E}{\mathbb E}

\newcommand{\Loss}{\mathcal{L}}
\newcommand{\LSTMVAE}{crVAE}

\newcommand{\LSTMVAEGAN}{crVAE-GAN}
\newcommand{\convVAE}{\text{cVAE}}
\newcommand{\blockVAE}{\text{bcVAE}}
\newcommand{\VAEGAN}{\text{VAE-GAN}}

\wacvfinalcopy 

\def\wacvPaperID{312} 

\ifwacvfinal\fi
\begin{document}

\title{Channel-Recurrent Autoencoding for Image Modeling}

\author{Wenling Shang \\
UvA-Bosch-Deltalab\\
{\tt\small wshang@uva.nl}
\and
Kihyuk Sohn \\
NEC Labs\\
{\tt\small ksohn@nec-labs.com}
\and
Yuandong Tian \\
Facebook AI Research\\
{\tt\small yuandong@fb.com}
}

\twocolumn[{%
\renewcommand\twocolumn[1][]{#1}%
\maketitle
\begin{center}
    \centering
   \includegraphics[width=1\textwidth]{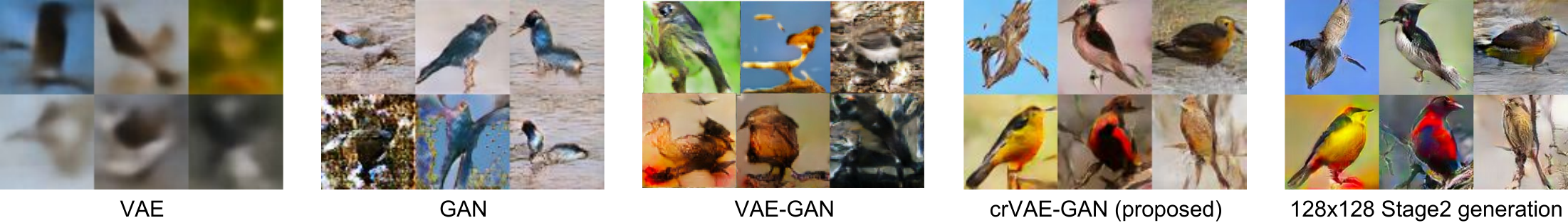}
    \captionof{figure}{Comparison demonstrating our channel-recurrent VAE-GAN's superior ability to model complex bird images. Based on the high-quality generation of Stage1 $64{\times}64$ images, higher-resolution Stage2 images can be further synthesized unsupervisedly.
}\label{fig:teaser}
\end{center}%
}]

\begin{abstract}
\cutabstractup
Despite recent successes in synthesizing faces and bedrooms, existing generative models struggle to capture more complex image types (Figure~\ref{fig:teaser}), potentially due to the oversimplification of their latent space constructions. 
To tackle this issue, building on Variational Autoencoders (VAEs), we integrate recurrent connections across \emph{channels} to both inference and generation steps, allowing the high-level features to be captured in global-to-local, coarse-to-fine manners. 
Combined with adversarial loss, our channel-recurrent VAE-GAN (crVAE-GAN) outperforms VAE-GAN in generating a diverse spectrum of high resolution images while maintaining the same level of computational efficacy. 
Our model produces interpretable and expressive latent representations to benefit downstream tasks such as image completion. 
Moreover, we propose two novel regularizations, namely the KL objective weighting scheme over time steps and mutual information maximization between transformed latent variables and the outputs, to enhance the training.
\end{abstract}
\begin{figure*}[t]
\centering
\subfigure[standard VAE]{\includegraphics[height=1.1in]{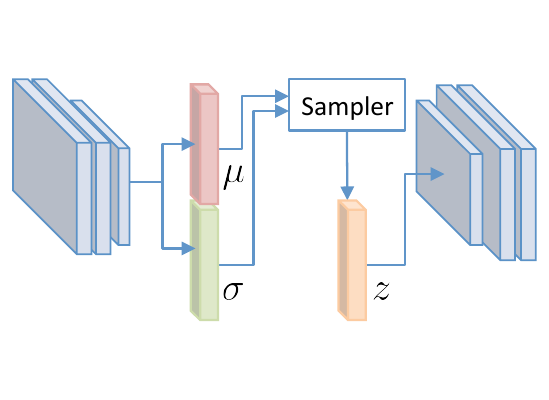}\label{fig:teasar_baseline}}
\hspace{0.04in}
\subfigure[convolutional VAE]{\includegraphics[height=1.1in]{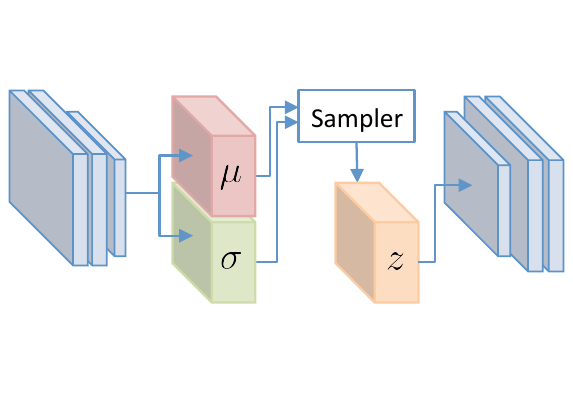}\label{fig:teasar_conv}}
\hspace{0.04in}
\subfigure[Proposed channel-recurrent VAE]{\includegraphics[height=1.1in]{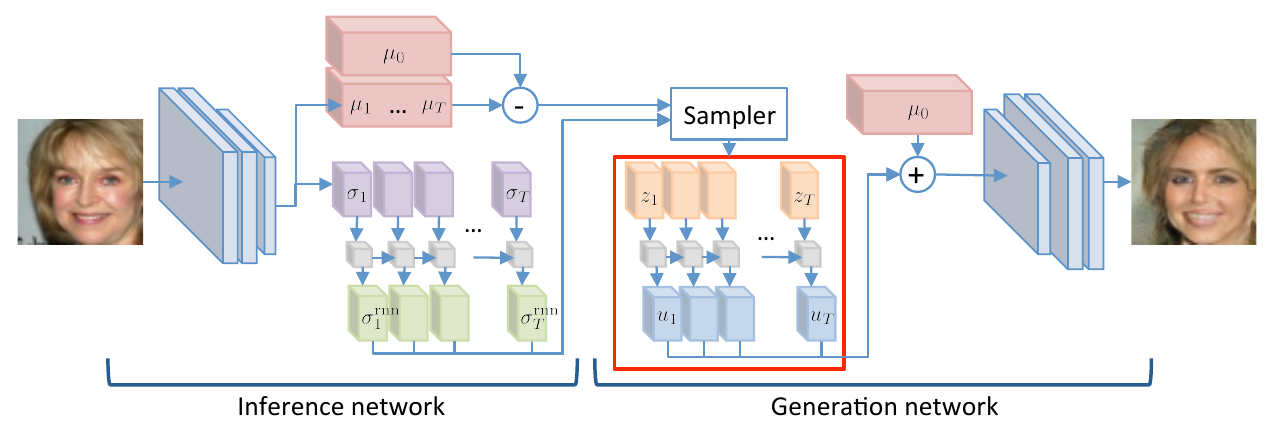}
\label{fig:teasar_lstm}}
\vspace{-0.15in}
\caption{Illustrations of (a) standard VAE, (b) its convolutional variant {\convVAE} and (c) the proposed {\LSTMVAE}.
}\label{fig:teasar}
\vspace{-0.23in}
\end{figure*}

\cutsectionup
\section{Introduction}
\label{sec:intro}
\cutsectiondown 
Tremendous progress has been made in generative image modeling in recent years.
Autogressive models, such as PixelRNN~\cite{oord2016conditional}, describe image densities autoregressively in the pixel level, leading to sharp generations albeit at a high computational cost and without providing latent representations.
Generative adversarial networks (GANs)~\cite{goodfellow2014generative} have shown promise, but are limited to modeling high density regions of data distributions~\cite{arora2017generalization,wu2016quantitative,theis2015note} and difficult to train~\cite{radford2015unsupervised,arjovsky2017wasserstein}.

Variational Autoencoder (VAE)~\cite{kingma2013auto} is a directed graphical model that approximates a data distribution through a variational lower bound (VLB) of its log likelihood.
VAE introduces an approximate posterior parameterized by a deep neural network (DNN) that probabilistically encodes the input to a latent representation. It is directly applicable for a wide range of downstream tasks from photo-editing~\cite{guo2017attribute} to policy learning~\cite{higgins2017darla}, which neither GANs nor autoregressive models are equipped with.
Inference in VAE can be done efficiently by a forward pass of the DNN, making it promising for real-time applications.
As opposed to GANs, the reconstruction objective of VAE assures a comprehensive input space mode coverage.
However, such wide mode coverage, when combined with the KL regularization to the approximated posterior, becomes a downside for modeling complex and high-dimensional images, resulting in blurry image generation~\cite{bousquet2017optimal}.
To resolve blurriness while preserving a meaningful latent space, \cite{larsen2015autoencoding} augments a VAE with an auxiliary adversarial loss, obtaining VAE-GAN.

However, recent works for high resolution ($64{\times}64$ and above) unsupervised image modeling are restricted to images such as faces and bedrooms, whose intrinsic degrees of freedom are low~\cite{larsen2015autoencoding,gans,mescheder2017adversarial}.
Once images become spatially and contextually complex, e.g. birds photographed in their natural habitats, aforementioned models struggle to produce sensible outputs (Figure~\ref{fig:teaser}), likely due to their latent space constructions lacking the capacity to represent complex input distributions.

In this work, we aim at resolving the limitations of VAE, such as blurry image generation and lack of expressiveness to model complex input spaces, in an unsupervised way.
While keeping graphical model unchanged to retain its original efficacy such as efficient probabilistic inference and generation, we propose to augment the architecture of inference and generation networks via recurrent connections across channels of convolutional features, leading to the \textit{channel-recurrent VAE} ({\LSTMVAE}).
Our approach is motivated by observing a common drawback to VAE and VAE-GAN: the fully-connected (FC) layers between the latent space and convolutional encoder/decoder. Although FC layers can extract abstract information for high-level tasks such as recognition~\cite{krizhevsky2012imagenet}, it omits much local descriptions that are essential for detailed image modeling as in our case.
Instead, we build latent features on convolutional activation without FC layers.
The proposed architecture sequentially feed groups of convolutional features sliced across channels into an LSTM~\cite{hochreiter1997long}, so that for each time step, the associated latent channels are processed based upon accumulated information from previous time steps to ensure temporal coherence while reducing the redundancy and rigidity from FC layers by representing distinguishing information at different time steps.
As a result, our model disentangles factors of variation by assigning general outlining to early time steps and refinements to later time steps. 
Analogously to VAE-GAN, We derive {\LSTMVAEGAN} by adding an additional adversarial loss, along with two novel regularization methods to further assist training.

We evaluate the performance of our {\LSTMVAEGAN} in generative image modeling of a variety of objects and scenes, namely birds~\cite{van2015building,berg2014birdsnap,wah2011caltech}, faces~\cite{liu2015deep}, and bedrooms~\cite{yu2015lsun}.
We demonstrate the superiority of our {\LSTMVAEGAN} qualitatively via $64{\times}64$ image generation, completion, and an analysis of semantic contents for blocks of latent channels, as well as quantitatively via inception scores~\cite{salimans2016improved} and human evaluation.
Specifically, significant visual enhancement is observed on the more spatially and contextually complex birds dataset.
We provide further empirical evidence through higher-resolution ($128{\times}128$ or $224{\times}224$) image synthesis by stacking an extra generation network on top of $64{\times}64$ generations from {\LSTMVAEGAN} and {\VAEGAN}, similarly to~\cite{zhang2016stackgan}. 
Unlike~\cite{zhang2016stackgan}, the success of generating higher-resolution 2nd stage images without condition variables is heavily dependent on the quality of the 1st stage generations. Our results verify the importance of channel-recurrent architecture in providing a solid 1st stage foundation to achieve high quality 2nd stage generation.
Lastly, we remark on the computational virtues of {\LSTMVAEGAN}.

The merits of our method are summarized as follows:
\vspace{-0.1in}
\begin{itemize}
\itemsep-0.5em
\item We integrate temporal structure to the latent space via LSTMs in replacement of the rigid FC layers to recurrently process latent channels, attaining a global-to-local, coarse-to-fine generation.
\item Our framework not only preserves the beneficial probabilistic latent space from VAE, allowing wide mode coverage, efficient posterior inference and training, but improves its expressiveness and interpretability. 
\item Our {\LSTMVAEGAN}, combined with two novel regularization methods, is capable of modeling complex input spaces when existing models fail. We visually and quantitatively demonstrate significant improvement on high-resolution image generation and related tasks over VAE-GAN. 
\item Our model, while producing state-of-the-art level image generations, maintains the computational efficacy from VAE. 
\end{itemize}
\vspace{-0.1in}
Code and pretrained models are published.

%
\cutsectionup
\section{Related Works}
\label{sec:related_work}
\cutsectiondown
\vspace{-0.03in}
Recent advances in deep generative modeling predominantly come from autoregressive models, Generative Adversarial Networks (GANs) and Variational Autoencoders (VAEs). 
Autoregressive models such as PixelRNN and PixelCNN~\cite{oord2016pixel,oord2016conditional,salimans2017pixelcnn++} directly characterize the probability density function over the pixel space. 
Although these models produce sharp images, they have slow inference, demand heavy GPU parallelization for training, and do not explicitly learn a latent space.
GAN~\cite{goodfellow2014generative} is another popular method in which a generator competes against a discriminator, producing outputs that imitate the inputs.
%
GANs suffer from several notable issues: limited distribution coverage~\cite{arora2017generalization,wu2016quantitative,theis2015note}, training instability~\cite{radford2015unsupervised,arjovsky2017wasserstein,mao2016multi} and lack of probabilistic latent spaces to encode a given input.  
VAEs~\cite{kingma2013auto} consist of a bottom-up inference network and a top-down generation network parameterized by DNNs that are jointly trained to maximize the VLB of the data log-likelihood. 
Although VAEs are mathematically elegant, easy to train, fast in inference and less GPU demanding than autoregressive models, its KL divergence penalty paired with reconstruction objective hampers realistic image generation since it overly stretches the latent space over the entire training set~\cite{theis2015note,bousquet2017optimal}.

Attempts are made to combine the aforementioned methods. 
PixelVAE~\cite{gulrajani2016pixelvae} integrates PixelCNN into VAE decoder but is still computationally heavy. 
RealNVP~\cite{dinh2016density} employs an invertible transformation between latent space and pixel space that enables exact log-likelihood computation and inference, but the model is restricted by the invertibility requirement. 
Adversarial Variational Bayes (AVB)~\cite{mescheder2017adversarial} theoretically builds more flexible approximated posterior via adversarial learning but empirically still outputs blurry generations.
VAE-GAN~\cite{larsen2015autoencoding} stitches VAE with GAN to enhance the generation quality while preserving an expressive latent space without introducing excessive computational overhead.
However, VAE-GANs are still not competitive in complex image classes as we note from Figure~\ref{fig:teaser}.
To tame complex input spaces, recent works~\cite{reed2016generative,yan2016attribute2image,odena2016conditional} leverage side information such as text description, foreground mask and class labels as conditional variables hoping that the consequential conditional distributions are less tangled.
But learning a conditional distribution requires additional labeling efforts both at training and downstream applications.
\begin{figure}[t]
\begin{center}
\centering
\subfigure[VAE]{\includegraphics[width=0.15\textwidth]{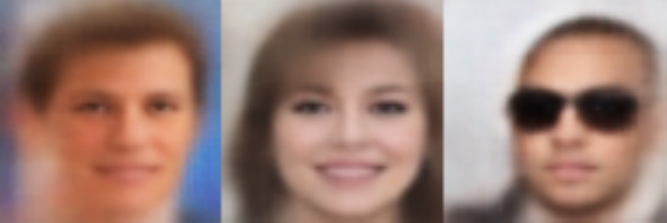}\label{fig:baseline_recon}} 
\subfigure[{\convVAE}]{\includegraphics[width=0.15\textwidth]{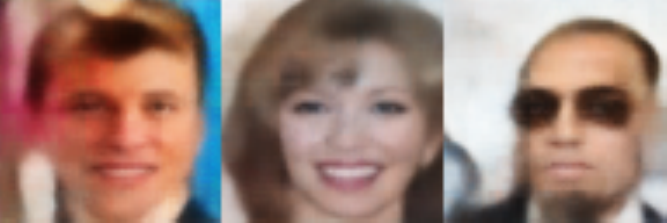}\label{fig:conv_recon}}
\subfigure[{\LSTMVAE}]{\includegraphics[width=0.15\textwidth]{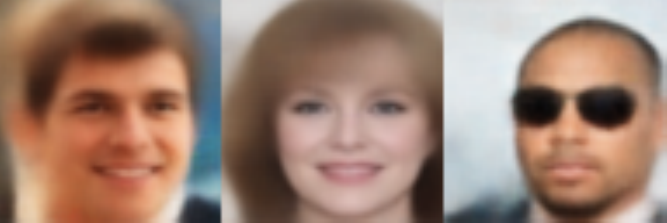}\label{fig:lstm_recon}}\\
\vspace{-0.1in}
\subfigure[VAE]{\includegraphics[width=0.15\textwidth]{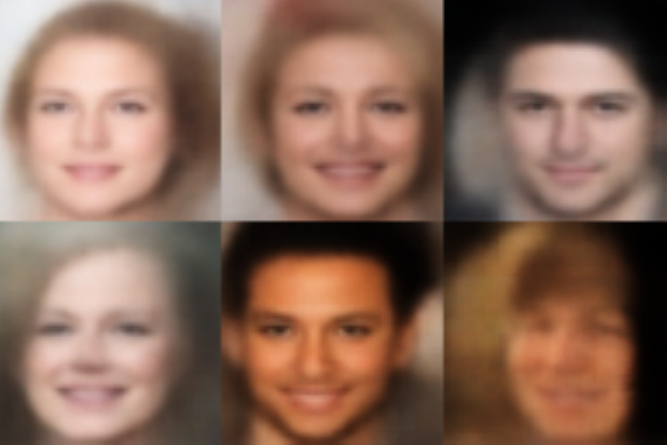}\label{fig:baseline_gen}}
\subfigure[{\convVAE}]{\includegraphics[width=0.15\textwidth]{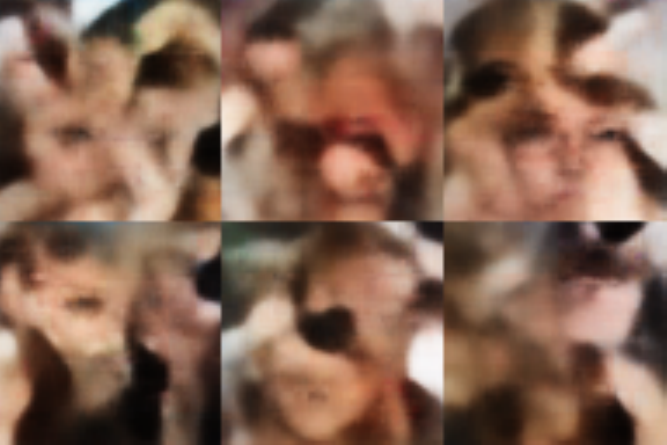}\label{fig:conv_gen}} 
\subfigure[{\LSTMVAE}]{\includegraphics[width=0.15\textwidth]{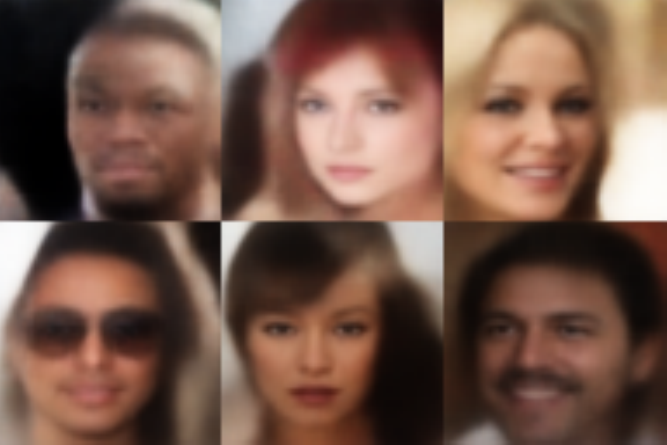}\label{fig:lstm_gen}}\\
\end{center}
\vspace{-0.3in}
\caption{(top) reconstructions with latent variables drawn from the approximated posterior and (bottom) generations from the prior for VAE, convolutional VAE ({\convVAE}), and the proposed {\LSTMVAE}.}
\label{fig:gens}
\vspace{-0.15in}
\end{figure}

Our approach handles complex input spaces well without the aid of conditional information.
It follows the pipeline of VAE-GAN but employs a core channel recurrency to transform convolutional features into and out of the latent space. 
Such changes are similar in spirit as recent works on improving approximate posterior and prior for VAEs~\cite{kingma2016improving,chen2016variational,rezende2015variational}, however, we do not change the prior or the posterior to maintain algorithmic simplicity and computational efficiency.
Our recurrent module builds lateral connections between latent channels following a similar philosophy as in the deep autoregressive networks (DARN)~\cite{gregor2014deep}.
But latent variables in DARN are sequentially drawn conditioned on the samples from the previous time steps and thus can be slow in inference.
DRAW networks~\cite{gregor2015draw,gregor2016towards} are related to ours as they also recurrently iterate over latent variables for generation. DRAW iterates over the entire latent variables multiple times and incrementally reconstructs pixel-level input at each iteration, whereas we only iterate between blocks of latent channels and reconstruct once, thus it is computationally more efficient and learns interpretable latent subspaces.
\cutsectionup
\section{Channel-Recurrent Autoencoding}
\label{sec:model_desc}
\cutsubsectionup
\vspace{-0.02in}
This section introduces the proposed channel-recurrent architecture, motivated by observing limitations of the standard VAE and convolutional VAE ({\convVAE}).
Then, we extend crVAE-GAN with an adversarial loss to render realistic images. 
Furthermore, we introduce two latent space regularization techniques specific to our proposed channel-recurrent architecture, namely, the KL objective weighting and the mutual information maximization.
\cutsubsectionup
\subsection{Latent Space Analysis of VAEs} 
\cutsubsectiondown
VAE approximates the intractable posterior of a directed graphical model with DNNs (Figure~\ref{fig:teasar_baseline}), maximizing a VLB of the data log likelihood:
\begin{equation}
\Loss_{\mathrm{VAE}} = -\E_{q_\phi (z|x)}\big[\log p_{\theta}(x|z)\big] + D_{\mathrm{KL}} ( q_\phi (z|x) \Vert p(z) ) \big]\nonumber
\end{equation}
where the approximate posterior $q_\phi (z|x) $ is modeled as a diagonal Gaussian and the prior $p(z)$ as a standard Gaussian.
We refer to $q_\phi (z|x)$ as an inference network and $p_{\theta}(x|z)$ a generation network.
The latent space of standard VAE is modeled as 1-dim vector $z{\in}\mathbb{R}^{c}$~\cite{kingma2013auto}, whereas for {\convVAE} the latent space is modeled as a 3-dim tensor $z{\in}\mathbb{R}^{w{\times}h{\times}c}$~\cite{sohn2015learning}. 

Overly smoothed reconstructions (Figure~\ref{fig:baseline_recon}) and generations (Figure~\ref{fig:baseline_gen}) as well as a lack of sample diversity are major downsides of VAEs.
A potential cause is that the naive parameterization of the latent space, its associated prior and approximated posterior, may not be able to reflect a complex data distribution~\cite{theis2015note,kingma2016improving,chen2016variational}.
One would be tempted to provide a fix by adding an image-specific prior to the latent space, such as spatial structure, leading to {\convVAE} (Figure~\ref{fig:teasar_conv}), whose inference and generation networks, different from standard VAE, are fully convolutional. 
By making the approximated posterior spatially correlated, {\convVAE} is able to learn with more local details during inference, reflected by higher quality reconstructions (Figure~\ref{fig:conv_recon}).
However, the latent variables sampled from spatially independent prior of {\convVAE} ignores the global structure of face shapes and produce chaotic samples (Figure~\ref{fig:conv_gen}). 
\begin{figure*}[t]
\centering
\subfigure[ground truth]{\includegraphics[width=0.27\textwidth]{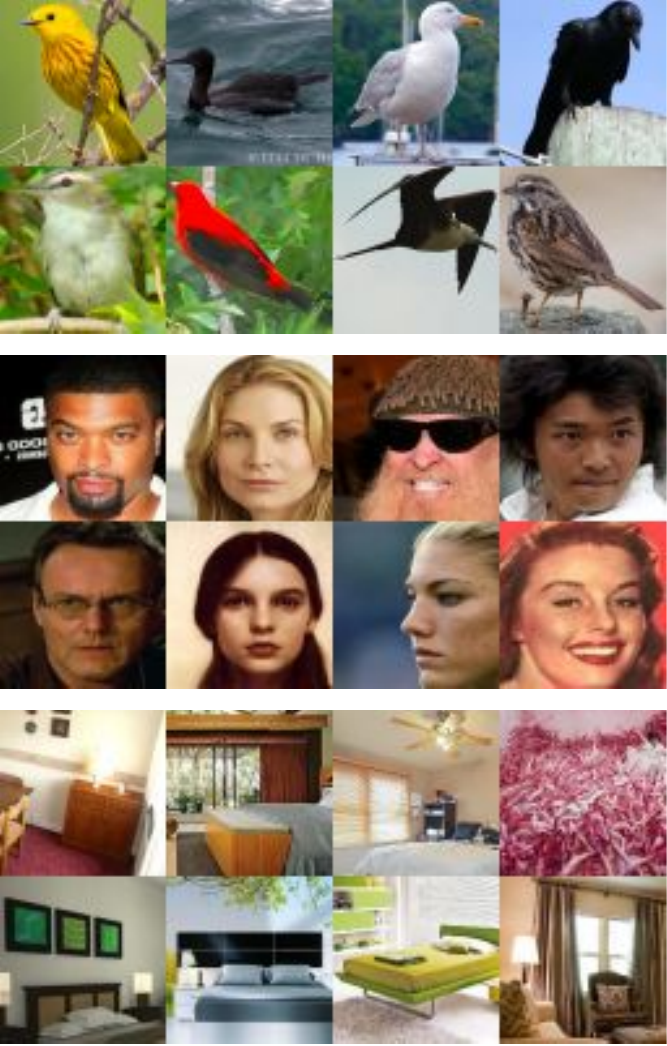}\label{64gt}}
\hspace{0.03in}
\subfigure[VAE-GAN]{\includegraphics[width=0.27\textwidth]{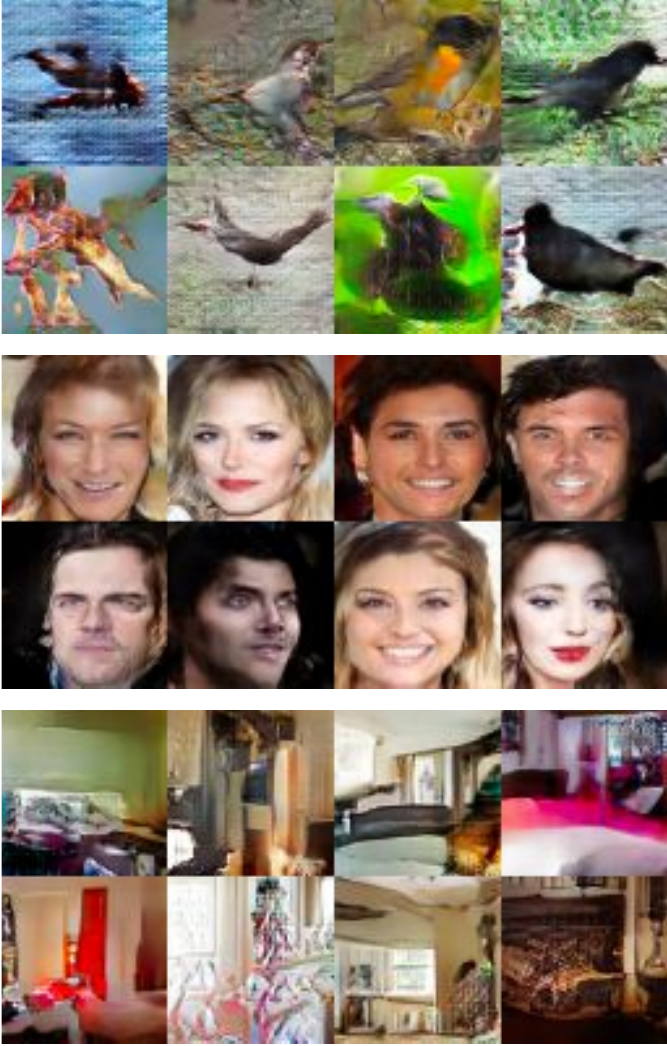}\label{64VAEGAN}}
\hspace{0.03in}
\subfigure[{\LSTMVAEGAN}]{\includegraphics[width=0.27\textwidth]{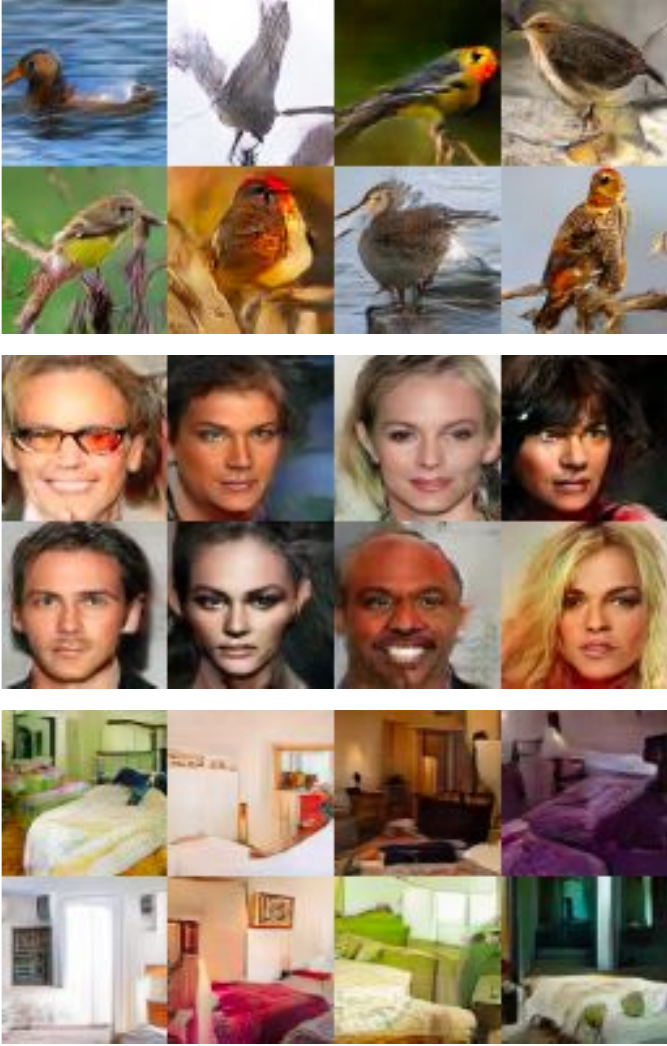}\label{64ours}}
\vspace{-0.15in}
\caption{64$\times$64 resolution image generation of (top) Birds, (middle) CelebA and (bottom) LSUN using (b) baseline VAE-GAN and (c) our proposed {\LSTMVAEGAN} along with (a) real examples.}
\label{fig:samples}
\vspace{-0.2in}
\end{figure*}
\cutsubsectionup
\subsection{Channel-Recurrent Variational Autoencoder} 
\cutsubsectiondown
Employing a prior with spatially dependent latent variables, such as a full-covariance Gaussian prior, is one remedy to the problem of chaotic image generation in {\convVAE s}.
However, such prior complicates the optimization due to significantly increased number of parameters (e.g., full covariance matrix $\sim O((w{\times}h{\times}c)^2)$), especially when the latent space is large~\cite{duchi2007derivations,gregor2014deep}.
Alternatively, we can structure the covariance to have dense dependencies across the spatial dimension and conditional dependencies along channels, in the hope that such setup can guide each channel to model different aspects of an image.
One possible way to apply the desired structure is to introduce hierarchy to the latent variables, which requires sequential sampling and complicates the training. Thus, tackling from a different direction, we opt to adapt the network architecture.
We propose to factorize the convolutional latent space into blocks across channels, flatten the activation to model spatial dependency, and connect the blocks via an LSTM~\cite{hochreiter1997long} to allow communication and coordination among them for coherency across channels.
That is, the transformation of the current block of latent channels always takes account of the accumulative information from the preceding time steps, serving as a guidance.
Concretely, during generation, $z {=} [z_1,\cdot\cdot,z_T]$ with $z_i{\in}\mathbb{R}^{w{\times}h{\times}\frac{c}{T}}$ sampled from standard Gaussian prior is passed through an LSTM to obtain a transformed representation $u = [u_1,\cdot\cdot,u_T] {=} \mathrm{LSTM}(z)$, which is then projected back to the pixel space.
Similarly, during inference, the mean path shares the same architecture as in {\convVAE}; the variance path slices latent variables into $T$ blocks of size $w{\times}h{\times}\frac{c}{T}$, each referred to as $\sigma_{i}$ and feeds $\sigma_{i}$'s into another LSTM to output $\sigma_{i}^{\mathrm{rnn}}$ as the final variances for the approximate posterior.
Our proposed model, referred to as the channel-recurrent VAE ({\LSTMVAE}), can both reconstruct and generate with higher visual quality than VAE and {\convVAE}, as shown in Figure~\ref{fig:lstm_recon} and~\ref{fig:lstm_gen}. 
More mathematical intuition and details for our design are in the Supplementary Materials.
\subsection{Additional Regularization}
\cutsectiondown
Inspired by~\cite{larsen2015autoencoding}, we adopt an adversarial loss to generate realistic images, leading to {\LSTMVAEGAN}.
Additionally, we propose two novel regularizers to enhance the latent space quality of {\LSTMVAEGAN} for better semantic disentanglement and more stable optimization.
\subsubsection{Generating Realistic Images with {\LSTMVAEGAN}}
\label{sec:crvae-gan}
We extend {\LSTMVAE} to {\LSTMVAEGAN} with an auxiliary adversarial loss on top of the generation network outputs for realistic image synthesis. The discriminator $D$ maps an image sampled from either the posterior or prior into a binary value:
\begin{gather}
\max_{\phi,\theta}\Loss_{\mathrm{VAE}} + \beta\E_{z\sim \{q_{\phi}(z|x),p(z)\}}\left[\log D(p_{\theta}(x|z))\right]\nonumber\\
\max_{D} \E_{x{\sim}X}\left[\log D(x)\right]{+}\E_{z{\sim}\{q_{\phi}(z|x),p(z)\}}\left[\log (1{-}D(p_{\theta}(x|z))\right].\nonumber
\end{gather}
Training can be done by min-max optimization as in~\cite{larsen2015autoencoding}.
\begin{figure}[t]
\begin{center}
\begin{minipage}{.5\textwidth}
\centering
\subfigure{\includegraphics[width=0.3\textwidth]{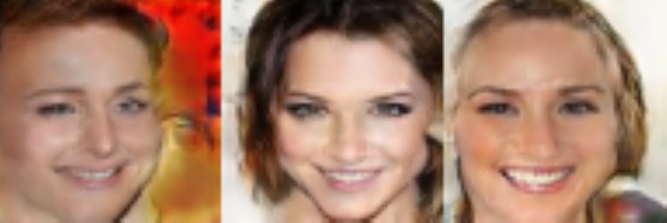}} 
\vspace{0.1in}
\subfigure{\includegraphics[width=0.3\textwidth]{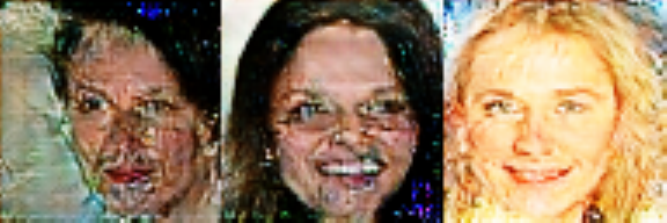}}
\subfigure{\includegraphics[width=0.3\textwidth]{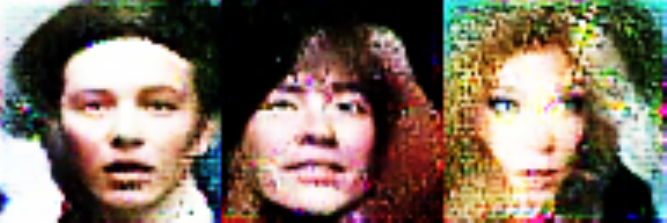}}
\end{minipage}
\vspace{-0.2in}
\end{center}
\vspace{-0.2in}
\caption{(Left) generations from the baseline VAE-GAN without MI regularization has much less artifacts than those from models trained with MI regularization on $z$ (middle) or $\mathrm{FC}_{\mathrm{gen}}(z)$ (right), implying such regularization is not compatible with VAE-GAN.}
\label{fig:baseline_MI}
\vspace{-0.2in}
\end{figure}
\subsubsection{Weighting the KL Objective} 
\begin{figure*}[t]
\centering
\includegraphics[width=0.95\textwidth]{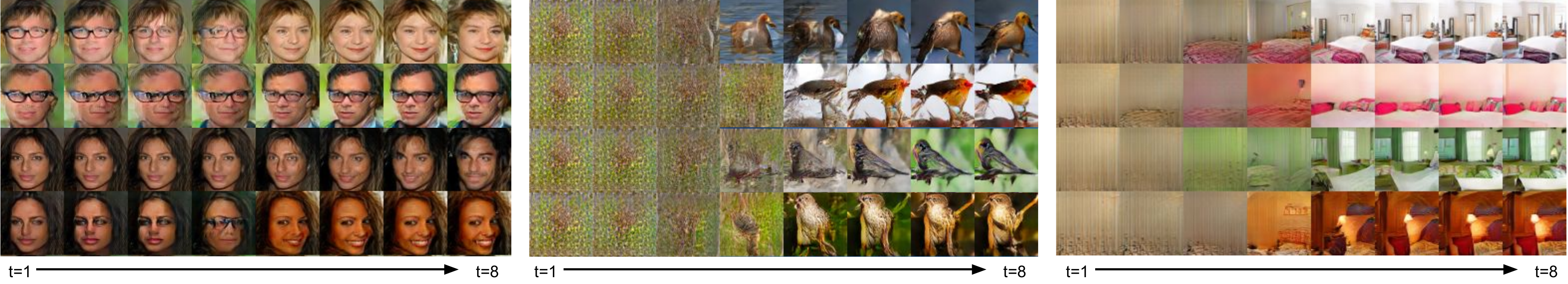}
\vspace{-0.15in}
\caption{Progressively drawing samples from a standard Gaussian prior over time steps. We observe how image generation evolves by determining global structure at earlier time steps and gradually adding more details later on. Recall the first 3 time steps carry more KL-weight than the rest hence a visual leap occurs around $t=3$ or $4$.}\label{fig:time_pro}
\vspace{-0.2in}
\end{figure*}

The KL objective of {\LSTMVAE} can be written as follows:
\begin{equation}
\sum_{t=1}^{T} (1-\alpha_{t})D_{\mathrm{KL}} ( q_\phi (z_{t}|x) \Vert p(z_{t}) ),
\end{equation}
where $\alpha_{t}{=}0,\;\forall t{\in}\{1,\cdots,T\}$ yields the equivalent expression to standard VAE objective.
The channel recurrent architecture additionally enables different weights to regularize the KL objective at each time step. 
Specifically, noting that the later time steps can be heavily influenced by the earlier ones due to the recurrent connection, we gradually reduce the regularization coefficients of the KL divergence to balance.
From an information theoretic perspective~\cite{alemi2016deep,tishby2000information}, the earlier time steps with larger coefficients hold a tighter information bottleneck, meaning that these latent channels shall convey general outlines, whereas the later time steps with smaller coefficients, i.e., a more flexible information flow, constitute diverse details conditioned on the sketched outlines. 
Figure~\ref{fig:time_pro} demonstrates the resulting effects where earlier time steps output rough profiles and later ones craft the details. 
\cutsubsubsectionup
\subsubsection{Mutual Information Regularization} 
\cutsubsubsectiondown
%
To increase training stability, we borrow the idea of mutual information (MI) maximization from~\cite{chen2016infogan} as an additional regularization. 
By recovering the latent variables from the generated image, the generation network encourages the administration of latent information to the output space. 
The regularization objective is written as:
%
\begin{gather}
\mathbb{E}_{z\sim \{q_{\phi}(z|x),p(z)\},\tilde{x}\sim p_{\theta}(x|z)}\big[q_{\psi}(z|\tilde{x})\big]\label{eq:MIreg}
\end{gather}
where $z$ can be sampled either from approximate posterior $q_{\phi}(z|x)$ or prior $p(z)$.
The CNN encoding path is shared between $q_{\psi}$ and $D$ as in~\cite{chen2016infogan}, mapping generated image to reconstruct $z$ and to a binary value, respectively.

Note that the formulation in Equation~\eqref{eq:MIreg} is not restricted to $z$, and we empirically found that relating the transformed representation $u{=}\mathrm{LSTM}_{\mathrm{gen}}(z)$ with the output of $q_{\psi}$ under {\LSTMVAEGAN} framework is much more effective.
However, similar regularization is found detrimental for {\VAEGAN}, both with $z$ and the transformed representation $\mathrm{FC}_{\mathrm{gen}}(z)$, as shown in Figure~\ref{fig:baseline_MI}. We compare samples generated from the baseline VAE-GAN and those trained with additional regularization to relate $z$ or $\mathrm{FC}_{\mathrm{gen}}(z)$ with the outputs, but the outcomes of the latter are visibly worse.
We contemplate that {\VAEGAN} lacks an equivalent transformation step as in {\LSTMVAEGAN} via channel-recurrent architecture, which particularly functions to enhance the latent representations.
More empirical demonstration are in Section~\ref{exp:MI} and implementation details in the Supplementary Materials.

\begin{figure*}[t]
\centering
\subfigure[{\LSTMVAEGAN} without MI regularization]{\includegraphics[width=0.49\textwidth]{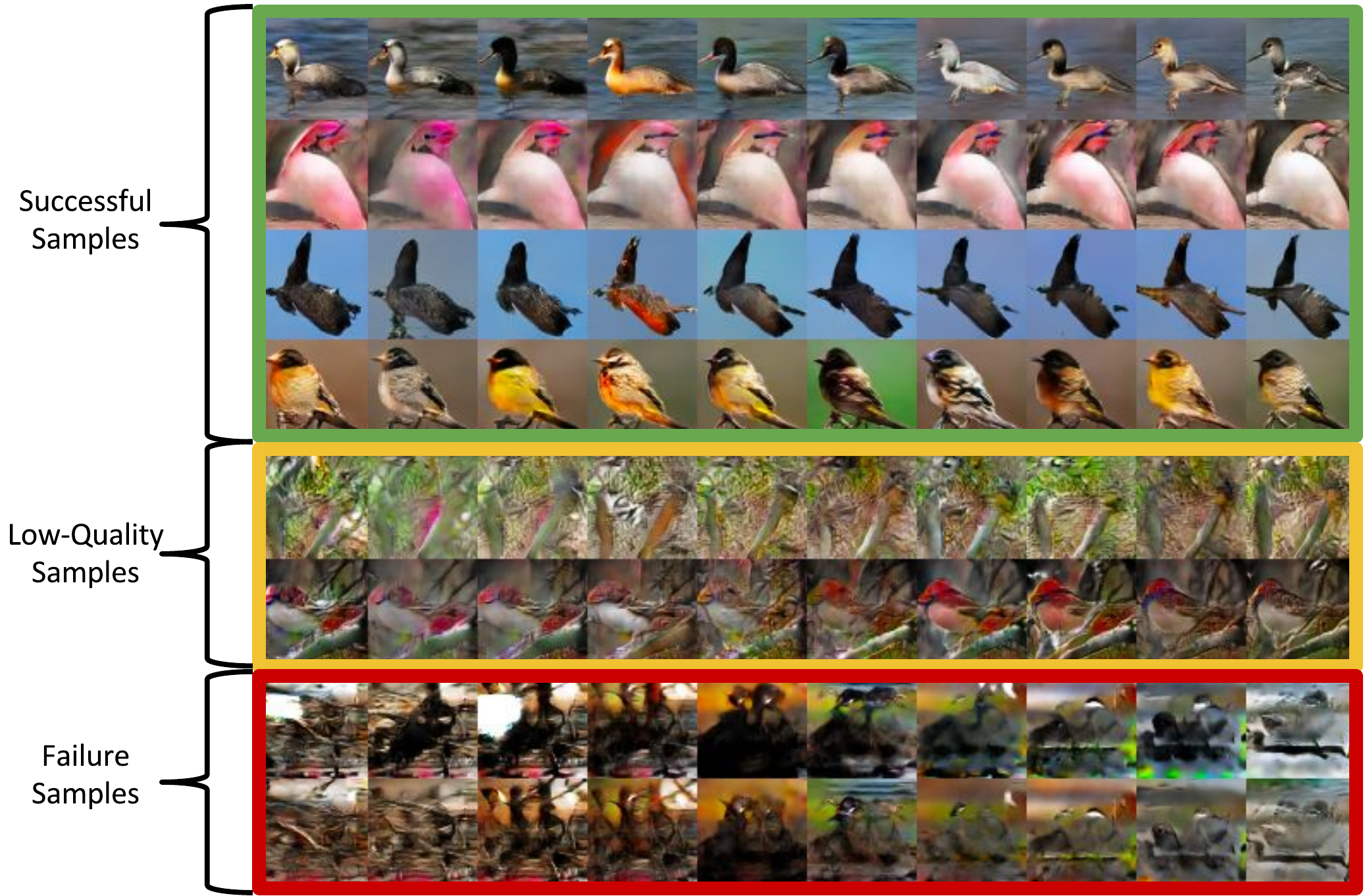}\label{fig:lstm-norecon}}\hspace{0.04in}
\subfigure[{\LSTMVAEGAN} with MI regularization]{\includegraphics[width=0.49\textwidth]{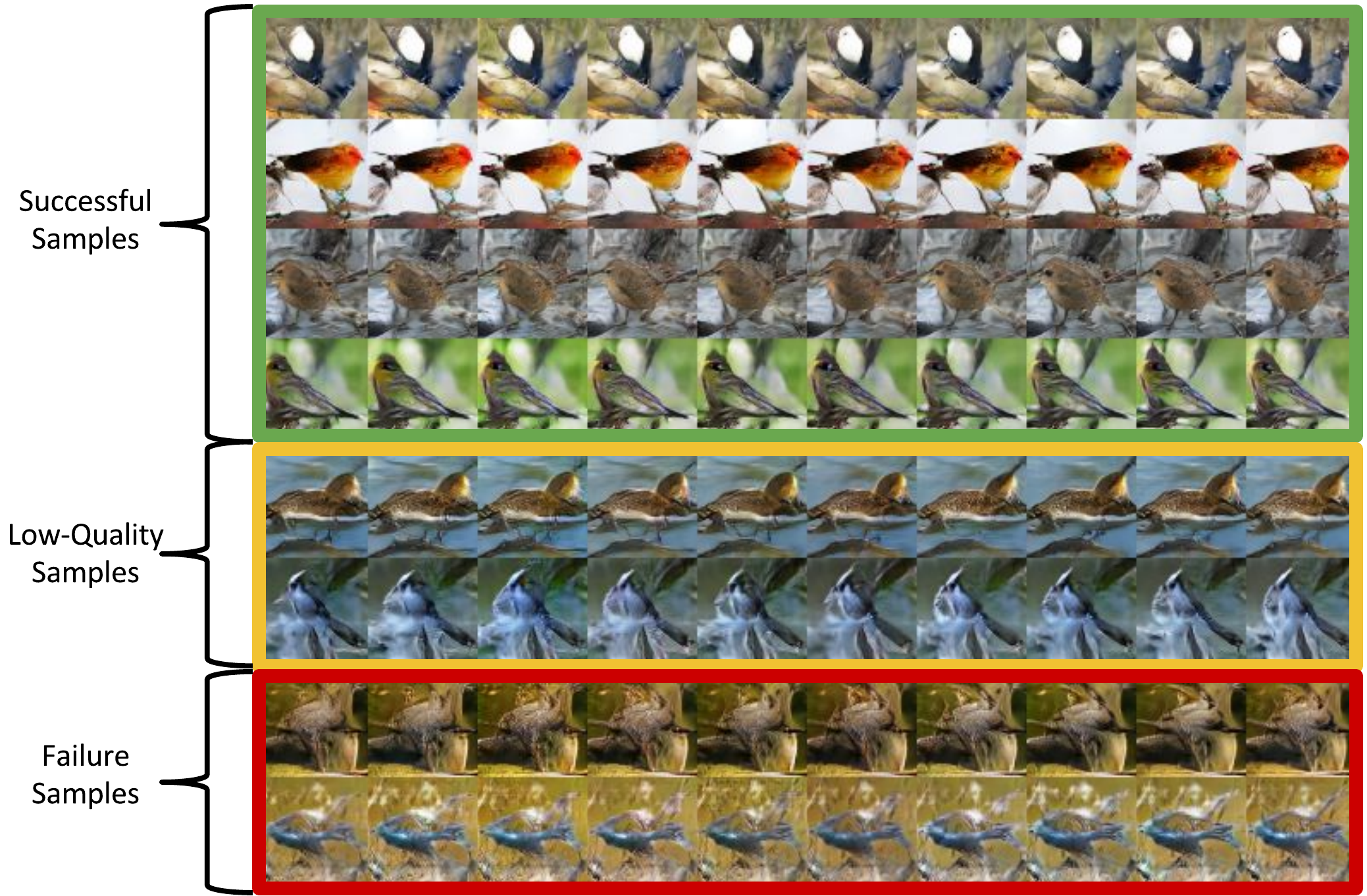}\label{fig:lstm-recon}}
\vspace{-0.15in}
\caption{The same $z$'s are sampled for each row from a standard Gaussian prior and projected back to the pixel space using the snapshot of decoders at last 10 epochs of training. (a) and (b) correspond to {\LSTMVAEGAN} trained without and with the MI regularization. Top 4 rows are successful samples, 5th and 6th are low-quality ones, and bottom 2 are failure cases. Clear improvement in stability is observed for (b) in terms of color oscillations between consecutive epochs and failure case mode collapsing.}
\vspace{-0.2in}
\label{fig:maxinfo}
\end{figure*}
\begin{figure*}[t]
\centering
\subfigure[CelebA: mouth, right-half, eyes]{\includegraphics[height=1.1in]{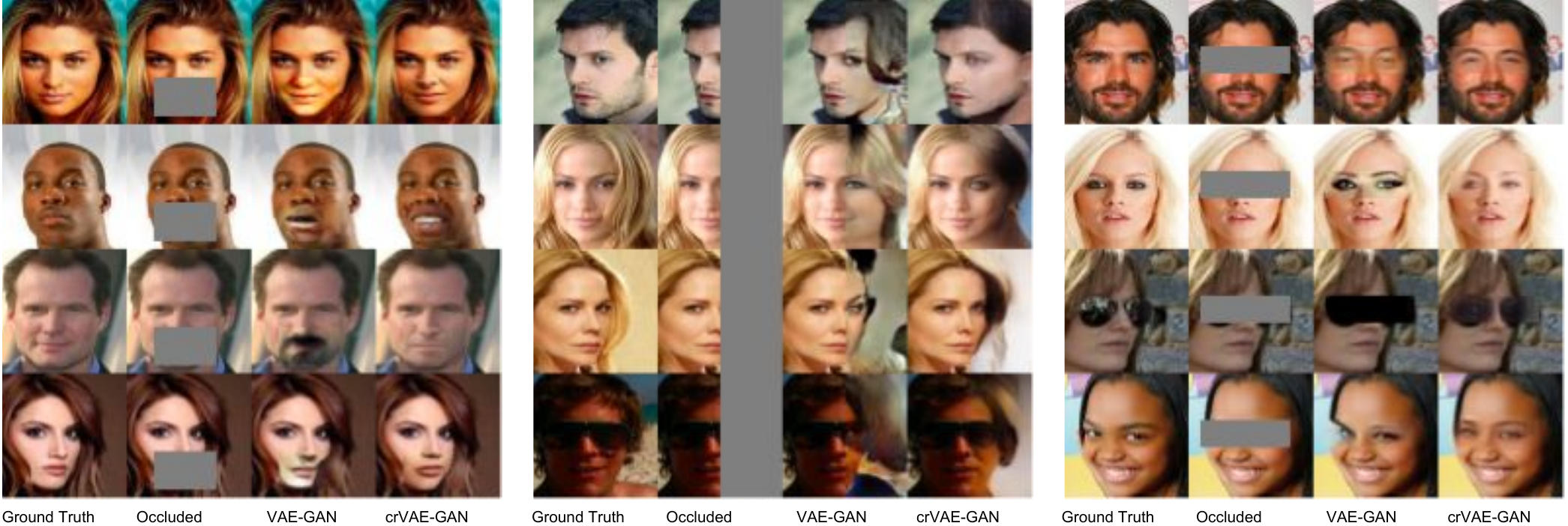}\label{fig:celeba_com}}\hspace{0.01in}
\subfigure[Birds: lower, upper]{\includegraphics[height=1.1in]{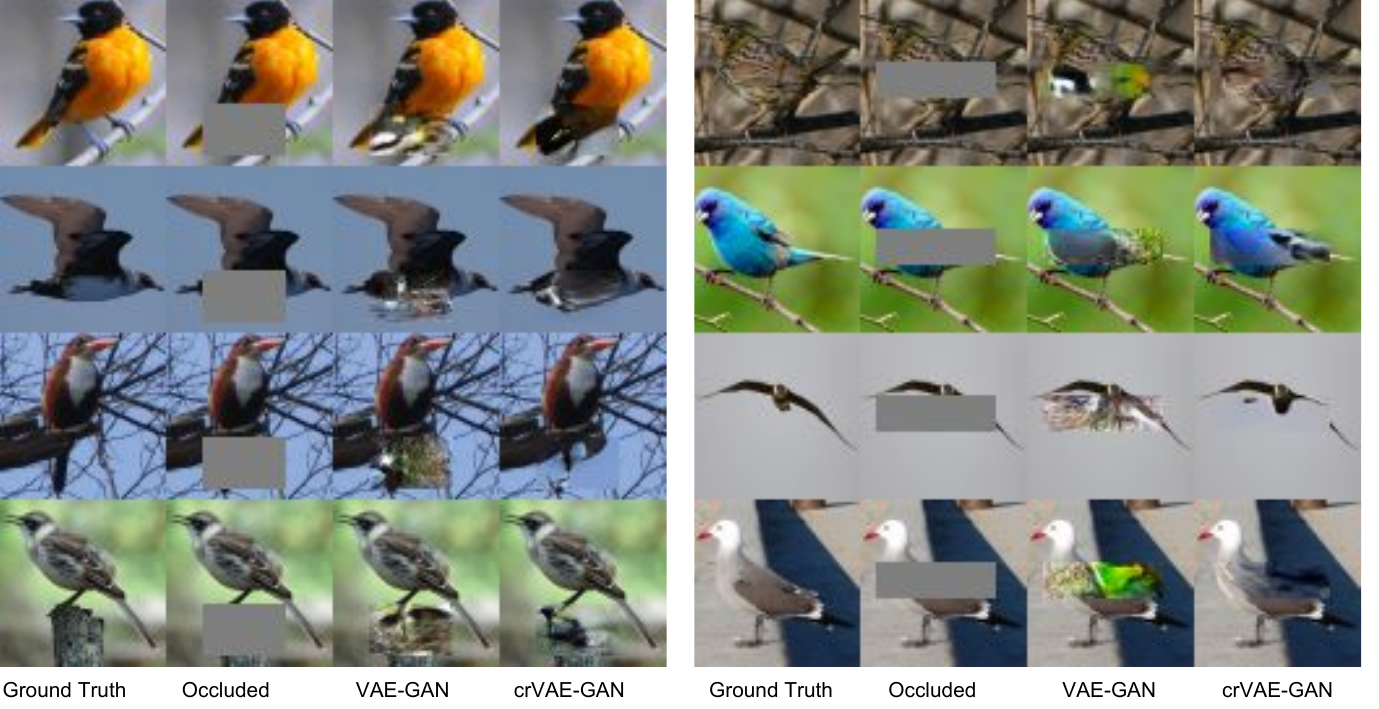}\label{fig:bird_com}}\hspace{0.01in}
\subfigure[LSUN: blocks]{\includegraphics[height=1.1in]{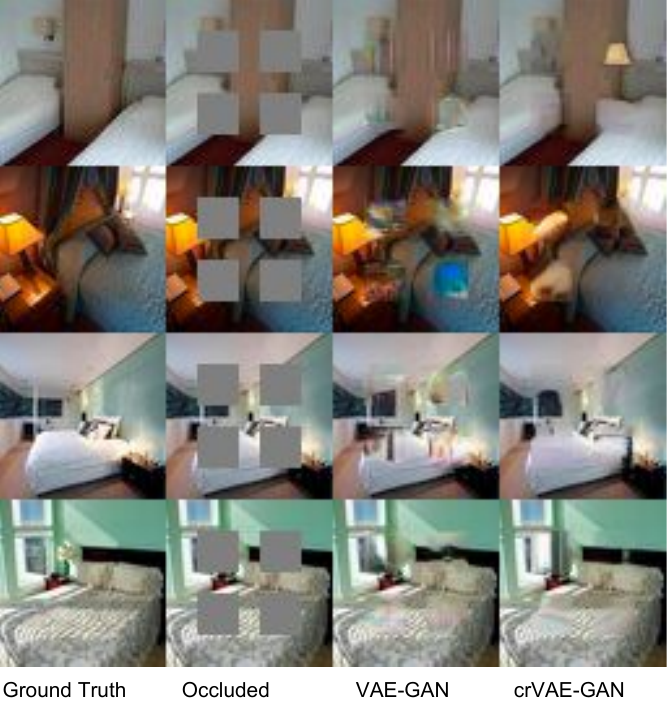}\label{fig:lsun_com}}\hspace{0.01in}
\vspace{-0.15in}
\caption{Image completion with {\VAEGAN} and {\LSTMVAEGAN}.
}
\vspace{-0.15in}
\label{fig:inpaint}
\end{figure*}
\cutsectionup
\section{Experiments}
\label{sec:experiments}
For evaluation, we first synthesize $64{\times}64$ natural images, followed by a 2nd stage generation to $128{\times}128$ or $224{\times}224$ on top of the 1st stage generation.
To demonstrate the latent space capacity,  image completion tasks are performed via optimization based on latent representations.
Finally, we explore the semantics of the learned latent channels, particularly with respect to different time steps.  
Three datasets, covering a diverse spectrum of contents, are used for evaluation.
Birds dataset is composed of three datasets, namely Birdsnap~\cite{berg2014birdsnap}, NABirds~\cite{van2015building} and Caltech-UCSD Birds-200-2011~\cite{wah2011caltech}, containing $106,474$ training and $5974$ validation images.
CelebA~\cite{liu2015deep} contains $163,770$ training and $19,867$ validation images of face.
LSUN bedroom (LSUN)~\cite{yu2015lsun} contains $3,033,042$ training and $300$ validation images. 
The ROIs are cropped and scaled to 64$\times$64 and 128$\times$128 for Birds and CelebA; the images are scaled and cropped to 64$\times$64 and 224$\times$224 for LSUN.  
Complete Implementation details are in the Supplementary Materials.

\begin{table}[t]
\centering
\small
\begin{tabular}{cccc}
\hline 
Model (64$\times$64) &  Birds &  CelebA &  LSUN \\ \hline
VAE-GAN &5.81\small{$\pm$0.09} & 21.70\small{$\pm$0.15} &    16.6$\%$        \\  \hline
{\LSTMVAEGAN} &10.62\small{$\pm$0.12} & 24.16\small{$\pm$0.33} & 29.9$\%$  \\  
{\LSTMVAEGAN} + MI &\textbf{11.07}\small{$\pm$0.12} &\textbf{26.20}\small{$\pm$0.19} & \textbf{53.5}$\%$   \\  \hline\hline
Model (128 or 224) &  Birds &  CelebA &  LSUN \\ \hline
VAE-GAN & 14.97\small{$\pm$0.11}&19.09\small{$\pm$0.19} &20.5$\%$\\   
VAE-GAN + Perc. & 14.61\small{$\pm$0.24}&27.09\small{$\pm$0.26}&10.5$\%$\\  \hline
{\LSTMVAEGAN}&29.14\small{$\pm$0.45}&\textbf{42.66}\small{$\pm$0.45} &\textbf{34.8}$\%$\\  
{\LSTMVAEGAN} + Perc. & \textbf{32.13}\small{$\pm$0.37} & 35.03\small{$\pm$0.39} &\textbf{34.2}$\%$\\   \hline
\end{tabular}
\vspace{0.05in}
\caption{Quantitative evaluation on generating 64$\times$64 and higher-resolution images. For Birds and CelebA, inception scores are reported. For LSUN, the frequency of selection by mechanical turk workers as the most realistc generation among all models is reported.}\label{tab:64}
\vspace{-0.3in}
\end{table}

\begin{table}[t]
\centering
\small
\begin{tabular}{ c  c  c  c }
\hline
 & Occlusion & {\VAEGAN} & {\LSTMVAEGAN}   \\  \hline
Birds & lower & $28.9\%$  & $\mathbf{71.1}\%$ \\ 
         & upper & $35.2\%$  & $\mathbf{64.8}\%$ \\ \hline
CelebA & eye & $18.0\%$ & $\mathbf{82.0}\%$ \\  
 & mouth & $22.7\%$ & $\mathbf{77.3}\%$ \\  
 & half & $34.4\%$ & $\mathbf{65.6}\%$ \\  \hline
LSUN & center & $23.4\%$ & $\mathbf{76.6}\%$ \\
\hline
\end{tabular}
\vspace{0.1in}
\caption{The frequency of selection by mechanical turk workers as the more realistic completion using {\VAEGAN} and {\LSTMVAEGAN}.}\label{tab:recon}
\vspace{-0.3in}
\end{table}
\subsection{Stage 1: 64$\times$64 Image Generation}
\label{sec:exp-stage1}
\cutsubsectiondown
We compare our {\LSTMVAEGAN} to the baseline VAE-GAN for 64$\times$64 image generation (Figure~\ref{fig:samples}). 
This also serves as the first step towards generating higher-resolution images later on.
For Birds, {\VAEGAN} generates colorful images (Figure~\ref{64VAEGAN}), but details are highly obscured, indicating the latent space conveys mostly low-level information such as color and edges, but not much high-level semantic concepts.
By contrast, {\LSTMVAEGAN} generates significantly more realistic birds with decent diversity in color, background and poses (Figure~\ref{64ours}).
Generating aligned faces and structured bedrooms are less difficult than birds, but {\LSTMVAEGAN} still exhibits clear superiority over {\VAEGAN}. 

For quantitative evaluation, we measure inception scores on Birds and CelebA using ImageNet pretrained VGG11 models finetuned on bird and face recognition task~\cite{yi2014learning}, respectively. 
As no bedroom classification dataset is available, we instead conduct a user study for LSUN. 
We ask a mechanical turk worker to select the most realistic out of 3 generated images (corresponding to VAE-GAN and {\LSTMVAEGAN}, with and without MI regularization), repeating for 2000 times.
We observe improved inception scores in Table~\ref{tab:64}, which agrees with the visual observation. 
The frequency of selection of each model by mechanical turk workers on generated LSUN images in Table~\ref{tab:64} also verifies that our proposed model outperforms the baseline with a significant margin. 
More non-curated image samples are in the Supplementary Materials.
\subsection{Effect of Mutual Information Regularization.}\label{exp:MI}
\cutsubsectiondown
We examine the effects of Mutual Information (MI) regularization on training {\LSTMVAEGAN} by projecting the same $z$ to the output pixel space via generation networks through the last 10 epochs of training. 
Figure~\ref{fig:lstm-norecon} and~\ref{fig:lstm-recon} show the results without and with MI regularization respectively, where top 4 rows are successful samples, 5th and 6th rows lower quality samples and bottom 2 rows failure cases. 
First, note that both models supply crisp, coherent and realistic samples when successful and their inception scores are also close (Table~\ref{tab:64}).
Nonetheless, without MI regularization, generated samples between consecutive epochs oscillate and failure cases tend to collapse to the same mode despite originating from different $z$'s. 
By contrast, with the regularization, the convergence becomes stable and the mode collapsing phenomenon no longer exists even for failed generations. 
High variance and mode collapsing are two well-known issues of adversarial training~\cite{radford2015unsupervised}. 
MI maximization overcomes these issues by (1) enforcing the latent messages to be passed to the outputs and (2) regulating the adversarial gradients--recall that MI and adversarial objectives share the same encoding path.
Unless specified otherwise, the {\LSTMVAEGAN} results reported are trained with MI regularization. 
\subsection{Stage2: Higher Resolution Image Generation}
\label{sec:exp-stage2}
\cutsubsectiondown
To further assess the quality of Stage1 64$\times$64 generations, we raise the generation resolution to 128$\times$128 for Birds and CelebA,\footnote{We decide to generate higher-resolution images of $128{\times}128$ for Birds and CelebA since the ROIs of are approximately of this resolution.} and to 224$\times$224 for LSUN. 
Our Stage2 generation network is designed similarly to that of StackGAN~\cite{zhang2016stackgan}, but generation is done in an unsupervised way without any condition variables. 
Thanks to the nature of our framework, the Stage1 outputs are composed of both generated and reconstructed samples.
We can utilize not only both sources of ``fake'' images in training but also an additional perceptual loss~\cite{johnson2016perceptual} of the reconstructed images to regularize the Stage2 network.
Figure~\ref{fig:up} presents generated samples from Stage2 networks with and without perceptual loss while taking generation outputs of {\VAEGAN} and {\LSTMVAEGAN} Stage1 models as input.
Table~\ref{tab:64} provides quantitative evaluations following the protocol of Section~\ref{sec:exp-stage1}.
The qualitative and quantitative results imply that a high quality Stage1 generation is essential for Stage2 success, despite that the Stage2 network can correct some of the Stage1 mistakes.
Since {\LSTMVAEGAN} supplies much higher quality Stage1 generations than {\VAEGAN}, it also produces more visually pleasing Stage2 generations.
%
We also observe that the inception scores in this case can diverge from visual fidelity, e.g. the Stage2 CelebA results with perceptual loss exhibit higher visual quality than without but lower inception scores.
Lastly, other mechanisms besides stacking generation networks can be applied to raise image resolution further such as variations of the recently proposed progressive GAN~\cite{progressive}, a worthy future direction but out of scope of this paper. 
More non-curated image samples, details on the Stage2 objectives and model architecture are in the Supplementary Materials.
\begin{figure*}[t]
\centering
\hspace{-0.02in}
\subfigure[expression ($t=8$)]{\includegraphics[height=0.55in]{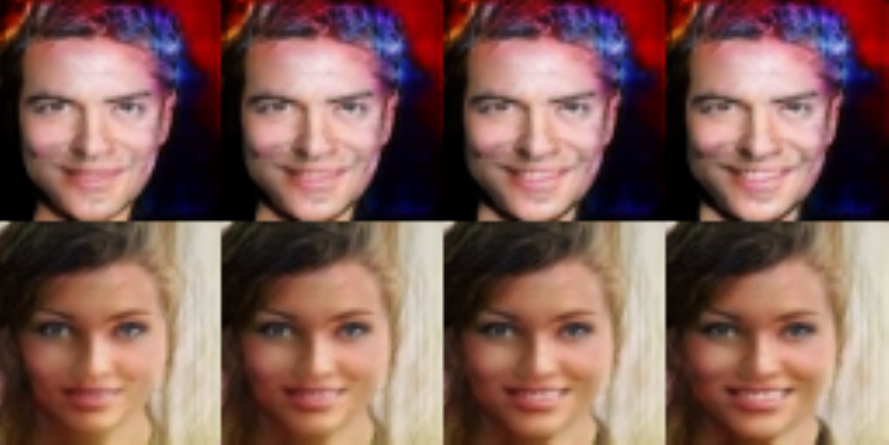}\label{fig:smile_pro}}
\hspace{-0.02in}
\subfigure[azimuth ($t=5$)]{\includegraphics[height=0.55in]{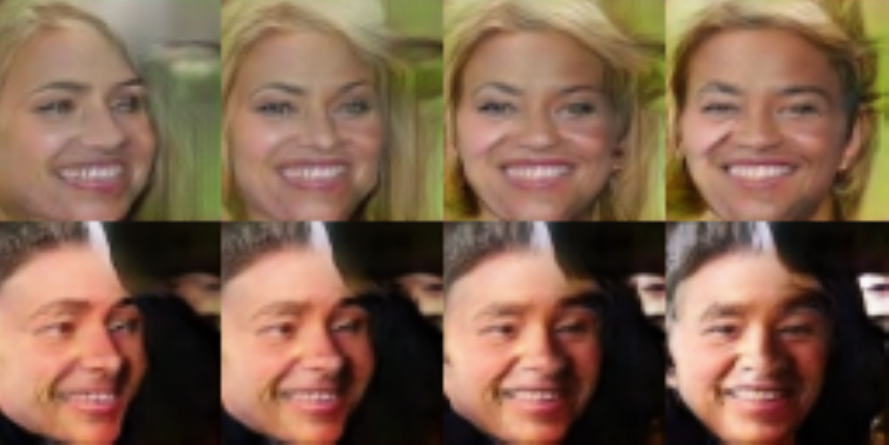}\label{fig:pose_pro}}
\hspace{-0.02in}
\subfigure[background ($t=5$)]{\includegraphics[height=0.55in]{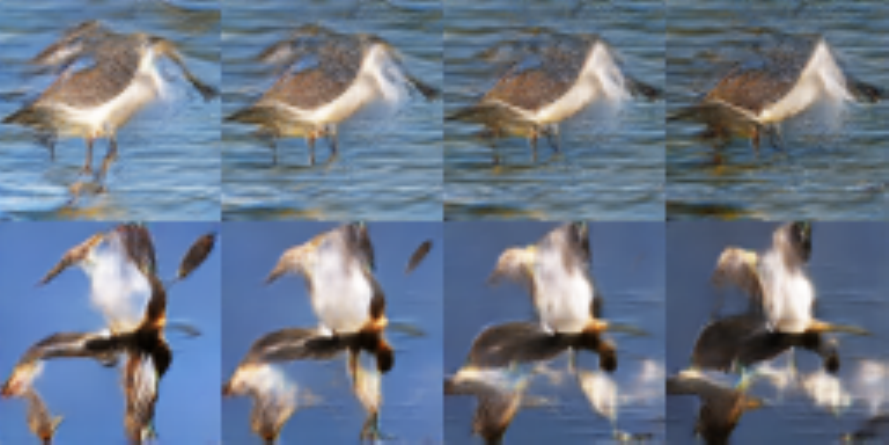}\label{fig:background_pro}}
\hspace{-0.02in}
\subfigure[belly size ($t=8$)]{\includegraphics[height=0.55in]{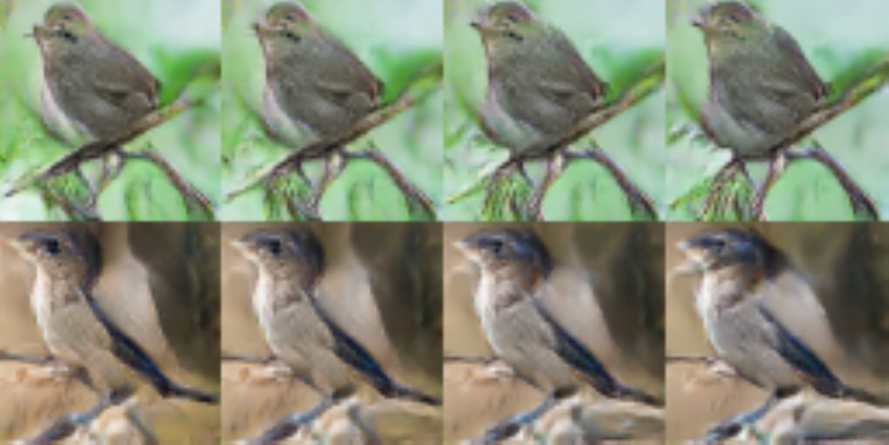}\label{fig:size_pro}}
\hspace{-0.02in}
\subfigure[color ($t=3$)]{\includegraphics[height=0.55in]{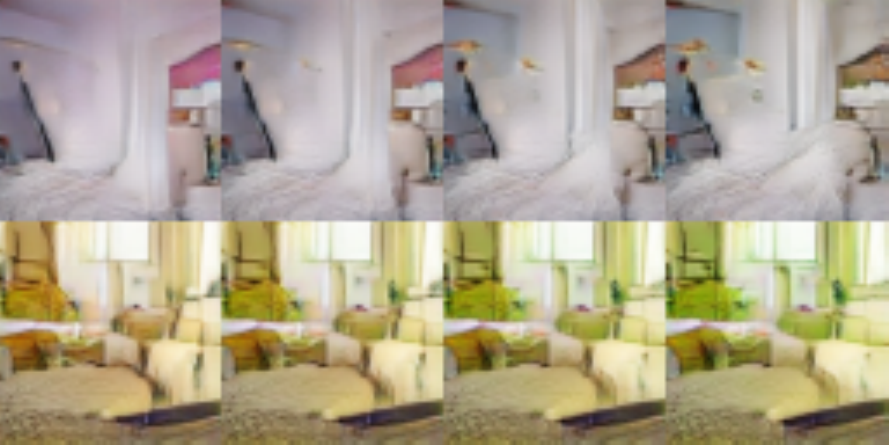}\label{fig:color_pro}}
\hspace{-0.02in}
\subfigure[window ($t=7$)]{\includegraphics[height=0.55in]{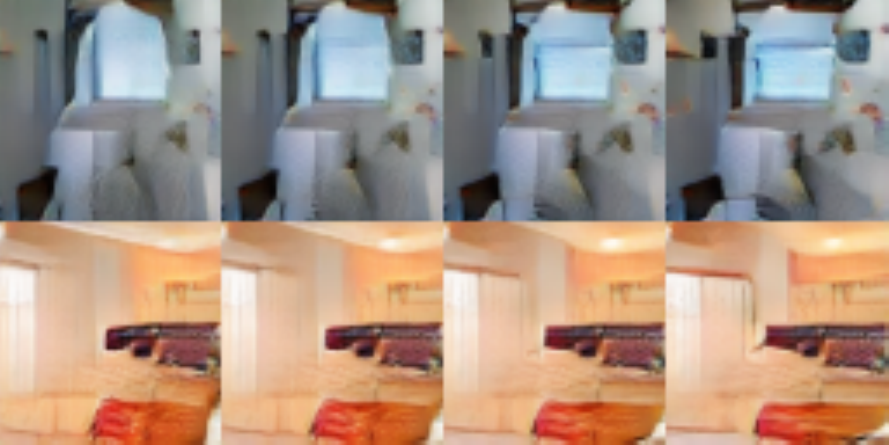}\label{fig:window_pro}}
\vspace{-0.15in}
\caption{Interpolating between $z_t$ and $\tilde{z}_t$, for a selected $t$, while fixing other $z_i$'s. We observe a gradual shift of an attribute towards the semantic direction encoded by $\tilde{z}_t$ while preserving most of the other factors.}
\label{fig:var}
\vspace{-0.2in}
\end{figure*}
\subsection{Image Completion}
%
To verify how faithfully the latent manifold from {\LSTMVAEGAN} reflects the semantic meaning of the input space, we conduct image completion task using Stage1 models.
We occlude parts of validation images, namely right-half, eye and mouth regions for CelebA, upper and lower part in Birds and blocks for LSUN, and then optimize their latent representations $z$'s to fill in the missing parts~\cite{yan2016attribute2image}: 
\begin{equation}
\min_z  \;\big[\Vert\hat{x}\odot m - x\odot m\Vert_{2}^{2} + \label{completion} \gamma \log \mathcal{N}(z;0,\mathbf{I}) + \tau \log (1-D(\hat{x})\big],\nonumber
\end{equation}
where $\hat{x}=\mathrm{gen}(z)$ refers to the output of the generation network, $m\in \{0,1\}^{3\times 64\times 64}$ is a mask whose entries are 0 if corresponding pixel locations are occluded and 1 otherwise, and $\odot$ represents element-wise multiplication.
Qualitative examples are in Figure~\ref{fig:inpaint}, with more in the Supplementary Materials. 
VAE-GAN struggles to complete some missing regions, e.g., right half of the faces and sunglasses in Figure~\ref{fig:celeba_com}, or generates excessive noise, e.g. the sky in Figure~\ref{fig:bird_com}. 
By contrast, {\LSTMVAEGAN} is more competent in retracting off-orbit latent points back to the actual manifold by better embedding the high-level semantic information.
Since there may exist multiple solutions in completing an image besides the ground truth, reconstruction error is not an ideal metric for quantitative measurement. Instead, we conduct human evaluation by presenting completed results from {\VAEGAN} and {\LSTMVAEGAN} to mechanical turk workers and asking them to select the more realistic one. 
The selection frequency of each model out of 128 randomly selected pairs is reported in Table~\ref{tab:recon}.
Overall our {\LSTMVAEGAN} again outperforms {\VAEGAN} with a substantial margin.

\subsection{Latent Channel Semantics}
\label{celeba:disentangle}
Our {\LSTMVAEGAN} processes the latent variables sequentially, allowing a global-to-local and coarse-to-fine progression.
Figure~\ref{fig:time_pro} highlights this progression by first initializing all latent variables to be zero, and then gradually sampling blocks of latent variables from the standard Gaussian prior.
Due to the weighted KL penalty, the first four time steps tend to operate on the overall tone of an image: defining background color theme, outlining the general shapes, etc. 
The second half attends the details: the texture of feathers for Birds, facial expressions for CelebA, lighting for LSUN, etc. 
For a concrete example, the first two rows from CelebA demonstration both starts with an outline of men with glasses, then gradually diverge to different hair styles, opposite poses, one taking off while the other solidifying the presence of glasses, and etc. 
Such progressing phenomenon also suggests that latent channels at different time steps carry their own interpretable semantics.

To investigate this hypothesis, we first draw random samples from the prior across 8 time steps, denoted by $z=[z_1, \cdot\cdot, z_t, \cdot\cdot, z_8]$; we then draw a new sample $\tilde{z} = [z_1, \cdot\cdot, \tilde{z}_{t}, \cdot\cdot, z_8]$ by only changing a representation at time $t$ that shows semantically meaningful changes from $z$; finally we generate images by interpolating between $z_t$ and $\tilde{z}_{t}$ while fixing other $z_i$'s. 
We note an interesting tendency where images sharing certain characteristics can be manipulated in the same way by interpolating towards the same $\tilde{z}_{t}$. 
For example, the CelebA samples in Figure~\ref{fig:time_pro} suggest that $t{=}5$ decides pose variation; indeed, if two faces both look into the right (Figure~\ref{fig:pose_pro}), traversing $z_5$'s of these faces to the same $\tilde{z}_{5}$ (selected by visual inspection) will smoothly frontalize their poses while retaining other factors.
%
Similar observations can also be made with Birds and LSUN.

Additionally, we conjecture that our model grants more freedom for semantic variations by associating an explainable factor to a latent subspace rather than a single latent unit~\cite{higgins2017beta,chen2016infogan}.
To investigate this hypothesis, we generate several different \emph{styles} of the same factor by sampling different $z_{t}$'s while fixing the other $z_i$'s.
For instance, Figure~\ref{fig:glasses} not only takes the glasses on/off but also switches from eyeglasses to sunglasses, from thin frame to thick frame.
In comparison, existing works on controlling factors of variation through latent unit manipulation, such as infoGAN~\cite{chen2016infogan} and $\beta$-VAE~\cite{higgins2017beta}, can only shift the controlled factor along a single direction, e.g. a latent unit that controls existence of glasses does not allow different glasses styles.

Channel recurrency is not yet perfect in explaining latent semantics. 
In particular, determining the direction representing a certain factor of variation still requires visual inspection.
Nevertheless, the preliminary demonstrations of this intriguing property shed light on future research in learning more semantically meaningful latent spaces.

\cutsectionup
\section{Computations}
\cutsectiondown
Another prominent advantage of {\LSTMVAEGAN} to the baseline as well as other state-of-the-art autoregressive models~\cite{oord2016conditional,kingma2016improving} is found from computational aspects.
First, as LSTMs share weights over time, for Stage1 generations, our proposed model has 130M parameters, when the baseline VAE-GAN with the same number of latent variables and the same encoder/decoder architecture has 164M. 
For the same reason, training our model consumes much less GPU memory. For example, in our implementation, during Stage1 optimization, {\LSTMVAEGAN} requires around 4.5GB memory with a batch size of 128 while {\VAEGAN} requires 6.2GB.
Finally, the inference and generation complexities for {\LSTMVAEGAN} is on the same order as those for {\VAEGAN}. 
In wall clock time, for a mini-batch of 128 images using a Titan X, {\LSTMVAEGAN} on average takes 5.8 ms for inference and 4.0 ms for generation; {\VAEGAN} takes 2.6 ms for inference and 2.2 ms for generation.
Meanwhile, autoregressive models are significantly slower in evaluation: even under careful parallelization, it is reported that PixelCNN~\cite{oord2016conditional} takes 52K ms and inverse autoregressive flow~\cite{kingma2016improving} 50 ms to generate a single 32$\times$32 image on a Titan X. 

\cutsectionup
\section{Conclusion}
\cutsectiondown
We propose the channel-recurrent autoencoding framework to improve the latent space constructions for image modeling upon the baseline VAE models. 
We evaluate the performance of our proposed framework via generative image modeling, such as image generation, completion, and latent space manipulation.
Future research includes building more interpretable features via channel recurrency and extrapolating our framework to other tasks.

\newpage
\subsubsection*{Acknowledgments}
We are grateful to Samuel Schulter, Paul Vernaza, Max Welling, Xiang Yu for their valuable comments on the manuscripts. We acknowledge Zeynep Akata for discussions during the preliminary stage of this work. We also thank NVIDIA for the donation of GPUs.

{\small
\bibliographystyle{ieee}
\bibliography{egbib}

\begin{thebibliography}{10}\itemsep=-1pt

\bibitem{alemi2016deep}
A.~A. Alemi, I.~Fischer, J.~V. Dillon, and K.~Murphy.
\newblock Deep variational information bottleneck.
\newblock {\em arXiv}, 2016.

\bibitem{arjovsky2017wasserstein}
M.~Arjovsky, S.~Chintala, and L.~Bottou.
\newblock Wasserstein gan.
\newblock {\em arXiv}, 2017.

\bibitem{arora2017generalization}
S.~Arora, R.~Ge, Y.~Liang, T.~Ma, and Y.~Zhang.
\newblock Generalization and equilibrium in generative adversarial nets (gans).
\newblock {\em ICML}, 2017.

\bibitem{berg2014birdsnap}
T.~Berg, J.~Liu, S.~Woo~Lee, M.~L. Alexander, D.~W. Jacobs, and P.~N.
  Belhumeur.
\newblock Birdsnap: Large-scale fine-grained visual categorization of birds.
\newblock In {\em CVPR}, 2014.

\bibitem{bousquet2017optimal}
O.~Bousquet, S.~Gelly, I.~Tolstikhin, C.-J. Simon-Gabriel, and B.~Schoelkopf.
\newblock From optimal transport to generative modeling: the vegan cookbook.
\newblock {\em arXiv}, 2017.

\bibitem{gans}
J.~Cha.
\newblock Implementations of (theoretical) generative adversarial networks and
  comparison without cherry-picking.
\newblock \url{https://github.com/khanrc/tf.gans-comparison}, 2017.

\bibitem{chen2016infogan}
X.~Chen, Y.~Duan, R.~Houthooft, J.~Schulman, I.~Sutskever, and P.~Abbeel.
\newblock Infogan: Interpretable representation learning by information
  maximizing generative adversarial nets.
\newblock In {\em NIPS}, 2016.

\bibitem{chen2016variational}
X.~Chen, D.~P. Kingma, T.~Salimans, Y.~Duan, P.~Dhariwal, J.~Schulman,
  I.~Sutskever, and P.~Abbeel.
\newblock Variational lossy autoencoder.
\newblock {\em arXiv}, 2016.

\bibitem{dinh2016density}
L.~Dinh, J.~Sohl-Dickstein, and S.~Bengio.
\newblock Density estimation using real nvp.
\newblock {\em ICLR}, 2017.

\bibitem{duchi2007derivations}
J.~Duchi.
\newblock Derivations for linear algebra and optimization.
\newblock {\em Berkeley, California}, 2007.

\bibitem{goodfellow2014generative}
I.~Goodfellow, J.~Pouget-Abadie, M.~Mirza, B.~Xu, D.~Warde-Farley, S.~Ozair,
  A.~Courville, and Y.~Bengio.
\newblock Generative adversarial nets.
\newblock In {\em NIPS}, 2014.

\bibitem{gregor2016towards}
K.~Gregor, F.~Besse, D.~J. Rezende, I.~Danihelka, and D.~Wierstra.
\newblock Towards conceptual compression.
\newblock In {\em NIPS}, 2016.

\bibitem{gregor2015draw}
K.~Gregor, I.~Danihelka, A.~Graves, D.~J. Rezende, and D.~Wierstra.
\newblock Draw: A recurrent neural network for image generation.
\newblock In {\em ICML}, 2015.

\bibitem{gregor2014deep}
K.~Gregor, I.~Danihelka, A.~Mnih, C.~Blundell, and D.~Wierstra.
\newblock Deep autoregressive networks.
\newblock In {\em ICML}, 2014.

\bibitem{gulrajani2016pixelvae}
I.~Gulrajani, K.~Kumar, F.~Ahmed, A.~A. Taiga, F.~Visin, D.~Vazquez, and
  A.~Courville.
\newblock Pixelvae: A latent variable model for natural images.
\newblock {\em arXiv}, 2016.

\bibitem{guo2017attribute}
Q.~Guo, C.~Zhu, Z.~Xia, Z.~Wang, and Y.~Liu.
\newblock Attribute-controlled face photo synthesis from simple line drawing.
\newblock {\em arXiv}, 2017.

\bibitem{higgins2017beta}
I.~Higgins, L.~Matthey, A.~Pal, C.~Burgess, X.~Glorot, M.~Botvinick,
  S.~Mohamed, and A.~Lerchner.
\newblock beta-vae: Learning basic visual concepts with a constrained
  variational framework.
\newblock In {\em ICLR}, 2017.

\bibitem{higgins2017darla}
I.~Higgins, A.~Pal, A.~A. Rusu, L.~Matthey, C.~P. Burgess, A.~Pritzel,
  M.~Botvinick, C.~Blundell, and A.~Lerchner.
\newblock Darla: Improving zero-shot transfer in reinforcement learning.
\newblock {\em ICML}, 2017.

\bibitem{hochreiter1997long}
S.~Hochreiter and J.~Schmidhuber.
\newblock Long short-term memory.
\newblock In {\em Neural Computation}, 1997.

\bibitem{johnson2016perceptual}
J.~Johnson, A.~Alahi, and L.~Fei-Fei.
\newblock Perceptual losses for real-time style transfer and super-resolution.
\newblock In {\em ECCV}, 2016.

\bibitem{progressive}
T.~Karras, T.~Aila, S.~Laine, and J.~Lehtinen.
\newblock Progressive growing of gans for improved quality, stability and
  variation.
\newblock {\em arXiv}, 2017.

\bibitem{kingma2016improving}
D.~Kingma, T.~Salimans, R.~Jozefowicz, X.~Chen, I.~Sutskever, and M.~Welling.
\newblock Improving variational inference with inverse autoregressive flow.
\newblock In {\em NIPS}, 2016.

\bibitem{kingma2013auto}
D.~Kingma and M.~Welling.
\newblock Auto-encoding variational bayes.
\newblock In {\em ICLR}, 2013.

\bibitem{krizhevsky2012imagenet}
A.~Krizhevsky, I.~Sutskever, and G.~E. Hinton.
\newblock Imagenet classification with deep convolutional neural networks.
\newblock In {\em NIPS}, 2012.

\bibitem{larsen2015autoencoding}
A.~B.~L. Larsen, S.~K. Sonderby, H.~Larochelle, and O.~Winther.
\newblock Autoencoding beyond pixels using a learned similarity metric.
\newblock In {\em ICML}, 2016.

\bibitem{liu2015deep}
Z.~Liu, P.~Luo, X.~Wang, and X.~Tang.
\newblock Deep learning face attributes in the wild.
\newblock In {\em ICCV}, 2015.

\bibitem{mao2016multi}
X.~Mao, Q.~Li, H.~Xie, R.~Y. Lau, and Z.~Wang.
\newblock Multi-class generative adversarial networks with the l2 loss
  function.
\newblock {\em arXiv}, 2016.

\bibitem{mescheder2017adversarial}
L.~Mescheder, S.~Nowozin, and A.~Geiger.
\newblock Adversarial variational bayes: Unifying variational autoencoders and
  generative adversarial networks.
\newblock {\em ICML}, 2017.

\bibitem{odena2016conditional}
A.~Odena, C.~Olah, and J.~Shlens.
\newblock Conditional image synthesis with auxiliary classifier gans.
\newblock {\em ICML}, 2017.

\bibitem{oord2016pixel}
A.~v.~d. Oord, N.~Kalchbrenner, and K.~Kavukcuoglu.
\newblock Pixel recurrent neural networks.
\newblock {\em ICML}, 2016.

\bibitem{oord2016conditional}
A.~v.~d. Oord, N.~Kalchbrenner, O.~Vinyals, L.~Espeholt, A.~Graves, and
  K.~Kavukcuoglu.
\newblock Conditional image generation with pixelcnn decoders.
\newblock {\em NIPS}, 2016.

\bibitem{radford2015unsupervised}
A.~Radford, L.~Metz, and S.~Chintala.
\newblock Unsupervised representation learning with deep convolutional
  generative adversarial networks.
\newblock {\em ICLR}, 2016.

\bibitem{reed2016generative}
S.~Reed, Z.~Akata, X.~Yan, L.~Logeswaran, B.~Schiele, and H.~Lee.
\newblock Generative adversarial text to image synthesis.
\newblock {\em ICML}, 2016.

\bibitem{rezende2015variational}
D.~Rezende and S.~Mohamed.
\newblock Variational inference with normalizing flows.
\newblock In {\em ICML}, 2015.

\bibitem{salimans2016improved}
T.~Salimans, I.~Goodfellow, W.~Zaremba, V.~Cheung, A.~Radford, and X.~Chen.
\newblock Improved techniques for training gans.
\newblock In {\em NIPS}, 2016.

\bibitem{salimans2017pixelcnn++}
T.~Salimans, A.~Karpathy, X.~Chen, and D.~P. Kingma.
\newblock Pixelcnn++: Improving the pixelcnn with discretized logistic mixture
  likelihood and other modifications.
\newblock {\em arXiv}, 2017.

\bibitem{sohn2015learning}
K.~Sohn, X.~Yan, and H.~Lee.
\newblock Learning structured output representation using deep conditional
  generative models.
\newblock In {\em NIPS}, 2015.

\bibitem{theis2015note}
L.~Theis, A.~v.~d. Oord, and M.~Bethge.
\newblock A note on the evaluation of generative models.
\newblock In {\em ICLR}, 2016.

\bibitem{tishby2000information}
N.~Tishby, F.~C. Pereira, and W.~Bialek.
\newblock The information bottleneck method.
\newblock {\em arXiv}, 2000.

\bibitem{van2015building}
G.~Van~Horn, S.~Branson, R.~Farrell, S.~Haber, J.~Barry, P.~Ipeirotis,
  P.~Perona, and S.~Belongie.
\newblock Building a bird recognition app and large scale dataset with citizen
  scientists: The fine print in fine-grained dataset collection.
\newblock In {\em CVPR}, 2015.

\bibitem{wah2011caltech}
C.~Wah, S.~Branson, P.~Welinder, P.~Perona, and S.~Belongie.
\newblock The caltech-ucsd birds-200-2011 dataset.
\newblock 2011.

\bibitem{wu2016quantitative}
Y.~Wu, Y.~Burda, R.~Salakhutdinov, and R.~Grosse.
\newblock On the quantitative analysis of decoder-based generative models.
\newblock {\em arXiv}, 2016.

\bibitem{yan2016attribute2image}
X.~Yan, J.~Yang, K.~Sohn, and H.~Lee.
\newblock Attribute2image: Conditional image generation from visual attributes.
\newblock In {\em ECCV}, 2016.

\bibitem{yi2014learning}
D.~Yi, Z.~Lei, S.~Liao, and S.~Z. Li.
\newblock Learning face representation from scratch.
\newblock {\em arXiv}, 2014.

\bibitem{yu2015lsun}
F.~Yu, A.~Seff, Y.~Zhang, S.~Song, T.~Funkhouser, and J.~Xiao.
\newblock Lsun: Construction of a large-scale image dataset using deep learning
  with humans in the loop.
\newblock {\em arXiv}, 2015.

\bibitem{zhang2016stackgan}
H.~Zhang, T.~Xu, H.~Li, S.~Zhang, X.~Huang, X.~Wang, and D.~Metaxas.
\newblock Stackgan: Text to photo-realistic image synthesis with stacked
  generative adversarial networks.
\newblock {\em arXiv}, 2016.

\end{thebibliography}
}


\end{document}